\documentclass[runningheads]{llncs}
\usepackage[T1]{fontenc}
\usepackage{graphicx}
\usepackage{hyperref}
\usepackage{color}
\usepackage[strings]{underscore}
\usepackage{booktabs}
\usepackage{float}
\usepackage{amsmath}
\usepackage{fancyvrb,fvextra}
\usepackage{subcaption}

\begin{document}

\title{Skill Learning Using Process Mining for Large Language Model Plan Generation}
\titlerunning{Skill Learning Using Process Mining for LLM Plan Generation}

\author{Andrei Cosmin Redis\inst{1} \and
Mohammadreza Fani Sani\inst{2} \and \\
Bahram Zarrin\inst{2} \and 
Andrea Burattin\inst{1}}
\authorrunning{AC Redis et al.}
\institute{Technical University of Denmark\\ \email{andreiredis@gmail.com} \\ \email{andbur@dtu.dk}\\ \and
Microsoft Development Center Copenhagen\\ 
\email{mfanisani@microsoft.com}\\
\email{bahram.zarrin@microsoft.com}}
\maketitle   
\vspace{-0.4cm}
\begin{abstract}

Large language models (LLMs) hold promise for generating plans for complex tasks, but their effectiveness is limited by sequential execution, lack of control flow models, and difficulties in skill retrieval. Addressing these issues is crucial for improving the efficiency and interpretability of plan generation as LLMs become more central to automation and decision-making. We introduce a novel approach to skill learning in LLMs by integrating process mining techniques, leveraging process discovery for skill acquisition, process models for skill storage, and conformance checking for skill retrieval. Our methods enhance text-based plan generation by enabling flexible skill discovery, parallel execution, and improved interpretability. Experimental results suggest the effectiveness of our approach, with our skill retrieval method surpassing state-of-the-art accuracy baselines under specific conditions.

\keywords{Large Language Model \and Plan Generation \and Process Mining \and Agentic Context Retrieval  \and Skill Learning}
\end{abstract}

%%%%%%%%%%%%%%%%%%%%%%%%%%%%%%%%%%%%%%%%%%%%%%%%%%%%%%%

\section{Introduction}

Large language models (LLMs) have demonstrated remarkable capabilities in natural language processing tasks and have shown some capacity for logical reasoning~\cite{wei_chain--thought_2024,kojima_large_2022}. However, their performance tends to decline as the complexity of problems increases, particularly in tasks requiring intricate reasoning or multi-step planning~\cite{gendron_large_2023,valmeekam_large_2023}. To enhance LLM performance on reasoning tasks, researchers have incorporated tools and plan generation capabilities, enabling LLMs to function as agents that generate sequences of tool invocations to solve given problems~\cite{ahn_as_2022,zhao_survey_2023,chang_survey_2023}.

Despite these advancements, existing plan generation methods, especially text-based planners, face significant challenges when dealing with complex tasks~\cite{valmeekam_large_2023}. Text-based planners typically produce flat sequences of actions without an underlying structured control flow model~\cite{zhao_survey_2023}. This lack of control flow limits their ability to generalize plans to other problems, making them less adaptable to varying parameters and conditions. Although these plans may include parameters, they are still less generalizable and often require adjustments or replanning when faced with new tasks. Additionally, the linear sequences are less interpretable, as the absence of logical structure makes it challenging for humans to understand the rationale behind the plans. Furthermore, the inability to identify parallelizable actions leads to less efficient execution, as tasks are processed sequentially without exploiting opportunities for concurrent execution.

In contrast, code-based planners~\cite{zhao_survey_2023} inherently incorporate structured control flow models, such as functions, loops, and conditionals, allowing for flexible and adaptable plan generation. This flexibility likely enables effective skill learning, as observed in works like Voyager by Wang et al.~\cite{wang_voyager_2023}, where previously generated plans (skills) can be reused and adapted to new problems. The structured control flow in code-based planners facilitates the grouping of related plans and the dynamic adjustment of actions based on different parameters and conditions.

The problem we address in this paper is the lack of structured control flow models in text-based LLM planners, which hinders their ability to perform efficient skill learning and limits their effectiveness in generating plans for complex tasks. Without a control flow model, text-based planners cannot effectively group related plans or identify parallelizable actions, resulting in sequential execution and increased latency.

To overcome this limitation, we propose a novel skill learning framework for text-based LLM planners that integrates process mining techniques~\cite{van_der_aalst_process_2016} to extract structured control flow models from flat action sequences. Process mining allows us to discover process models from execution traces, providing a structured representation of the control flow underlying the sequences generated by text-based planners. By incorporating these process models, we enable text-based planners to benefit from structured control flow, similar to code-based planners, thereby enhancing skill learning and plan generation capabilities.

Our approach addresses the limitations of text-based planners by:

\begin{enumerate}
    \item \textbf{Enabling flexible skill discovery and storage}: By discovering process models from action sequences, we can capture general behaviors and reuse skills across similar problems, reducing the reliance on generating plans from scratch.
    \item \textbf{Supporting parallel execution of actions}: The structured control flow models identify ordering constraints and parallelizable tasks, allowing for parallel execution where appropriate, thus reducing execution time and service latency.\addtocounter{footnote}{-2}\footnote{For example, in the TaskBench dataset~\cite{shen_taskbench_2023}, enabling parallel execution could potentially answer queries 1.43 times faster, assuming equal execution time for all actions. The time to execute a process model where groups of actions without ordering constraints can be executed in parallel is equal to the longest path of the plan in process model format (e.g., BPMN).}
    \item \textbf{Improving interpretability and reliability}: Structured process models could enhance the interpretability of the LLM's decision-making process, facilitating debugging and optimization by developers and users.
\end{enumerate}

\begin{figure}[tb]
    \centering
    \vspace{-.5cm}
    \includegraphics[width=\linewidth]{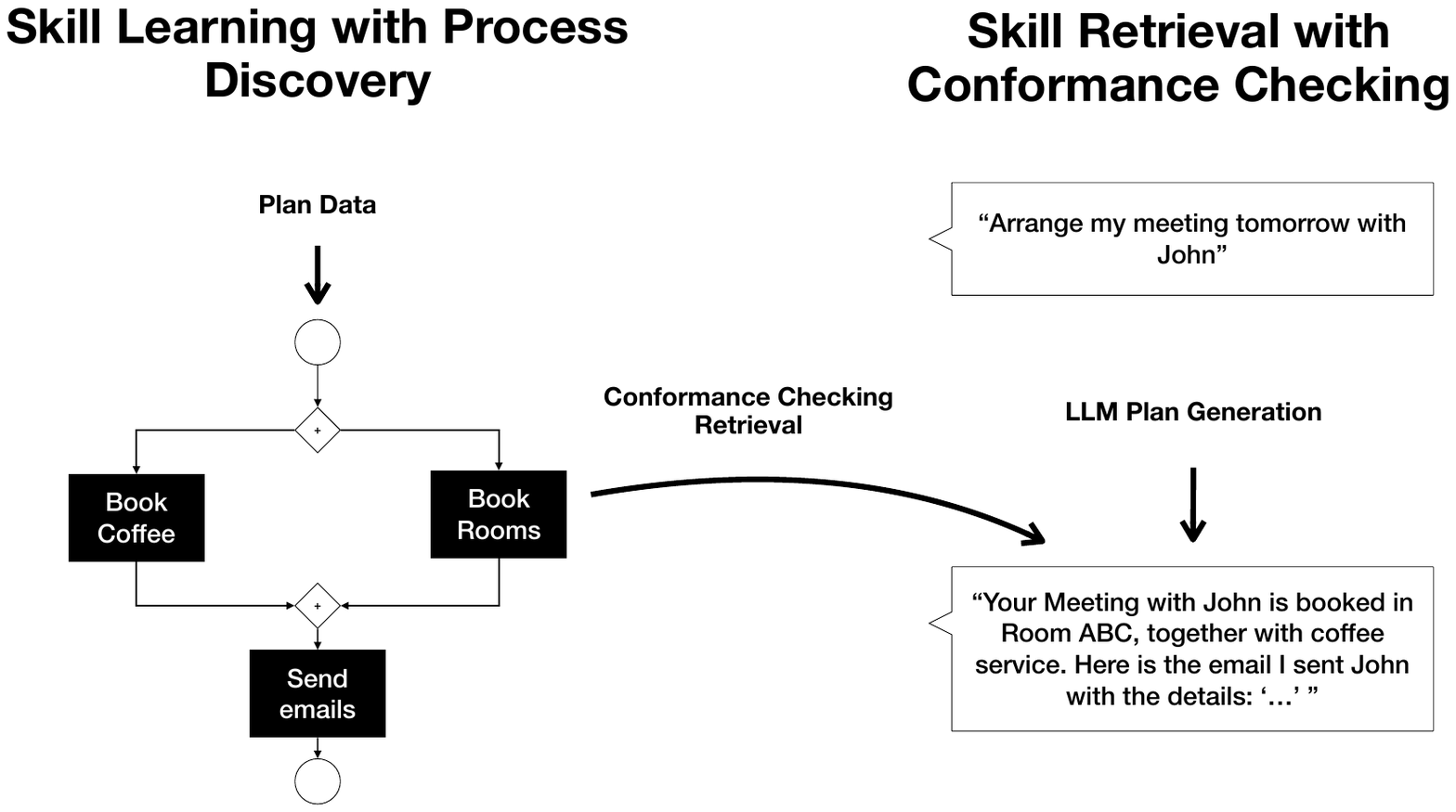}
    \vspace{-1cm}
    \caption{In the skill learning approach, when given a prompt such as 'Arrange my meeting tomorrow with John,' the LLM plan generator retrieves the 'meeting' skill to enhance its response. This paper introduces process mining techniques to discover this skill in a process model format, facilitating its retrieval and offering additional advantages, such as enabling parallelism in plan execution.}
    \label{fig-over}
    \vspace{-.5cm}
\end{figure}

As illustrated in Fig.~\ref{fig-over}, when given a prompt such as "Arrange my meeting tomorrow with John," the LLM plan generator can retrieve the "meeting" skill represented as a process model. This additional context enhances its response by leveraging the structured control flow and enabling parallel execution where appropriate.

Throughout this paper, we will use the terms \textit{plans}, \textit{traces}, and \textit{cases} interchangeably. Similarly, the terms \textit{process}, \textit{problem}, \textit{query}, and \textit{task}. That is also the case for \textit{action}, \textit{activity}, and \textit{step}, as well as in equal measure \textit{planner}, \textit{agent}, and \textit{LLM plan generator}. This is to reflect on the close relationship between the fields of planning and process mining and the overlap in the concepts they deal with.

%%%%%%%%%%%%%%%%%%%%%%%%%%%%%%%%%%%%%%%%%%%%%%%%%%%%%%
\vspace{-0.2cm}
\section{Related Work}

In addition to the related work discussed in the introduction, several other studies intersect with the field of process mining. Fettke et al. corroborate our findings from developing our approach, asserting that despite the compartmentalized research, AI planning, machine learning, and process mining share common objectives, making collaboration advantageous ~\cite{fettke_towards_2022}. Moreover,~\cite{fani2023llms} explores the potential of utilizing LLMs for event log abstraction and process automation.

Regarding process mining and LLMs, research has primarily concentrated on employing LLMs to perform process mining tasks. Despite the existing research into prompt engineering for process mining ~\cite{jessen_chit-chat_nodate}, process question answering~\cite{de_weerdt_abstractions_2024}, and event log data pre-processing~\cite{de_weerdt_large_2024}, the application of process mining methods to assist with LLM tasks remains largely uncharted.

%%%%%%%%%%%%%%%%%%%%%%%%%%%%%%%%%%%%%%%%%%%%%%%%%%%%%%%
% \vspace{-0.2cm}
\section{Skill Learning Using Process Discovery}

Text-based LLM planners typically generate flat sequences of actions to solve tasks. While effective for simple scenarios, these linear plans lack the structured control flow needed for complex tasks involving parallelism and reusability. To address this limitation, we propose a process mining method to discover structured control flow models —  \textit{skills} — from these flat action sequences.

Our approach takes the action sequences generated by the LLM planner as input and applies process discovery techniques, such as the Inductive Miner algorithm~\cite{leemans_partial-order-based_2023}, to infer general process models. These models capture the underlying control flow, including sequential and parallel relations between actions, providing a more expressive and flexible representation than flat plans.

Storing these skills as process models in a skill library enable the LLM planner to reuse previous solutions when faced with similar problems, reducing the need to generate new plans from scratch. This approach is analogous to the skill learning observed in code-based planners like Voyager~\cite{wang_voyager_2023} but applied to text-based planners without requiring them to generate code.

\begin{figure}[th]
    \centering
    \vspace{-1.6cm}
    \includegraphics[width=\linewidth]{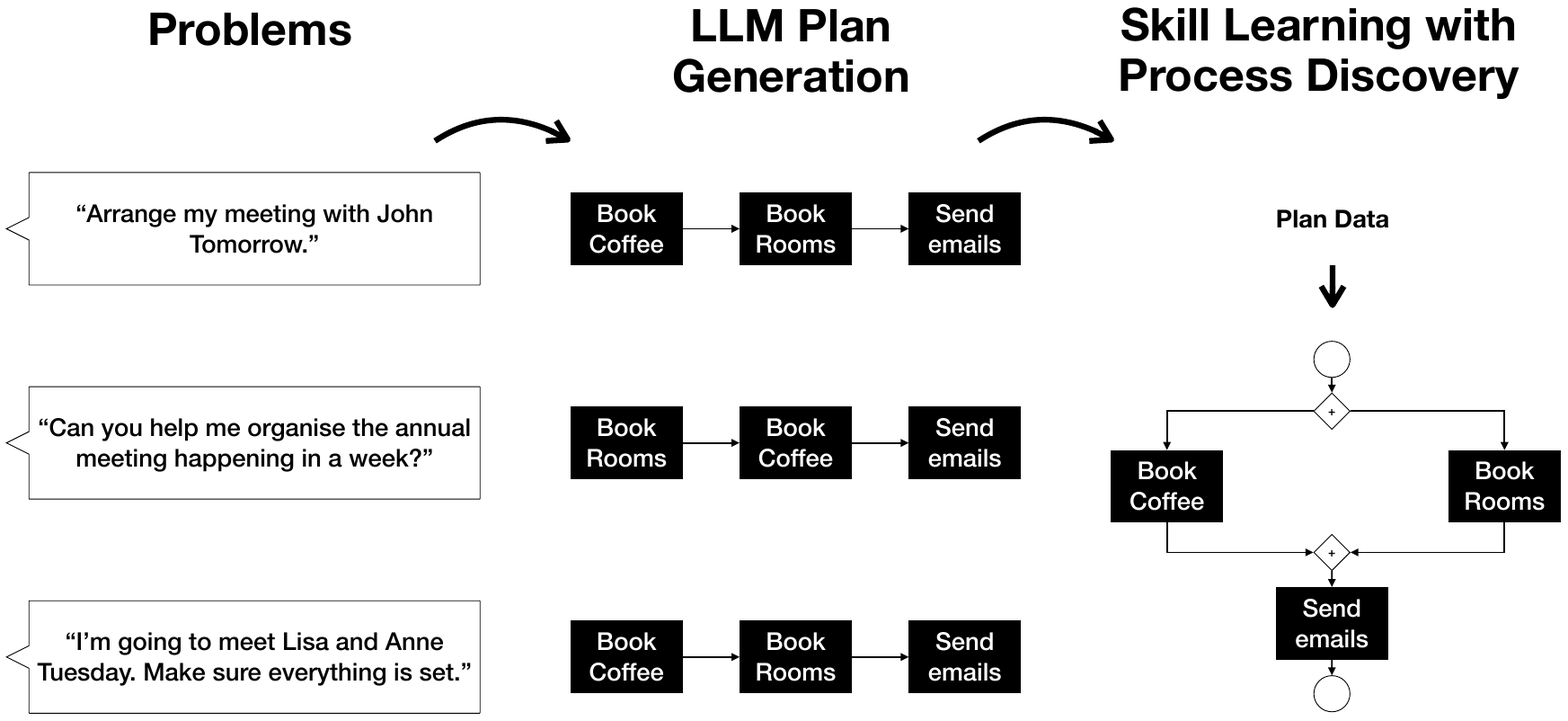}
    \vspace{-1.45cm}
    \caption{Schematic view of the skill learning process using process discovery.}
    \label{fig-discovery}
    \vspace{-0.3cm}
\end{figure}

The skill learning framework overview is shown in Fig.~\ref{fig-discovery}. The inputs are problems, and their corresponding flat plans are generated by the LLM. We transform these plans into structured process models (skills) through process mining, which are then stored in the skill library for future retrieval.

By integrating process mining into the skill learning process, our method enhances text-based LLM planners by providing them with reusable, interpretable, and parallelizable skills derived from their own generated plans. This bridges the gap between text-based and code-based planners, enabling text-based planners to handle more complex tasks effectively while maintaining their simplicity.

\section{Skill Retrieval Using Conformance Checking}
\label{sec-methods-retrieval}

To enhance both the accuracy and interpretability of skill retrieval in LLMs, we propose two methods based on conformance checking: (1) retrieval solely using conformance checking and (2) a two-stage retrieval method that combines text semantic similarity with conformance checking. The overall pipeline is shown in Figure~\ref{fig-over}.

\begin{enumerate}
    \item Conformance Checking Only: This method exclusively relies on conformance checking. This process mining technique assesses how well the control flow of a candidate skill's process model aligns with a bare-bones LLM-generated plan, referred to as a \textit{thought}. We define \textit{thought} as a partial plan embodying a full planning trace but not needing to be fully grounded in the state space. I.e., a plan that does not consider information received from the execution of tools, such as would be the case with \cite{yao_react_2023}. The key metric is alignment fitness, which measures the degree of structural match between the generated plan and the stored process models. Focusing on structural alignment rather than textual similarity offers superior interpretability, making it easier to understand why a particular skill was retrieved. The direct comparison of control flows ensures that the retrieved skills are relevant and logically compatible with the problem at hand.
    \item Two-Stage Retrieval: This hybrid method begins with a rapid filtering stage using text semantic similarity. Embedding models like the Universal Sentence Encoder (USE) or OpenAI's ada-002 generate vector representations of the problem description, and cosine similarity is used to identify the top-k candidates. These shortlisted candidates are then reranked using conformance checking based on alignment fitness. This two-stage approach balances computational efficiency with high retrieval accuracy and enhanced interpretability, as the final ranking is based on the logical structure of the process models rather than solely on textual descriptions.
\end{enumerate}

Both methods aim to outperform existing baselines that rely primarily on semantic similarity between text descriptions. By integrating process model alignment, these methods improve the precision of skill retrieval and significantly enhance the retrieval process's interpretability. This ensures that the logic behind the skill selection is transparent, which is crucial for debugging and optimizing LLM-driven plan executions. Additionally, the alignment values obtained from conformance checking can help assess the quality of the generated plan.
 
%%%%%%%%%%%%%%%%%%%%%%%%%%%%%%%%%%%%%%%%%%%%%%%%%%%%%%%
% \vspace{-0.4cm}
\section{Evaluation and Discussion}

In this section, we aim to answer the following questions: Given the current state of LLM plan generation, how feasible is it to learn skills using our proposed skill learning method? And, enabled by the skills learned, how does the accuracy of the proposed skill retrieval methods compare with previous approaches?

\subsection{Experiments}
\label{sec-experiments}

\renewcommand{\thesubfigure}{~\thefigure\alph{subfigure}:}
\captionsetup[subfigure]{labelformat=empty}

\begin{figure}[H]
    \centering
    \vspace{-0.7cm}
    \begin{subfigure}[t]{0.65\linewidth}
        \centering
        \includegraphics[width=\linewidth]{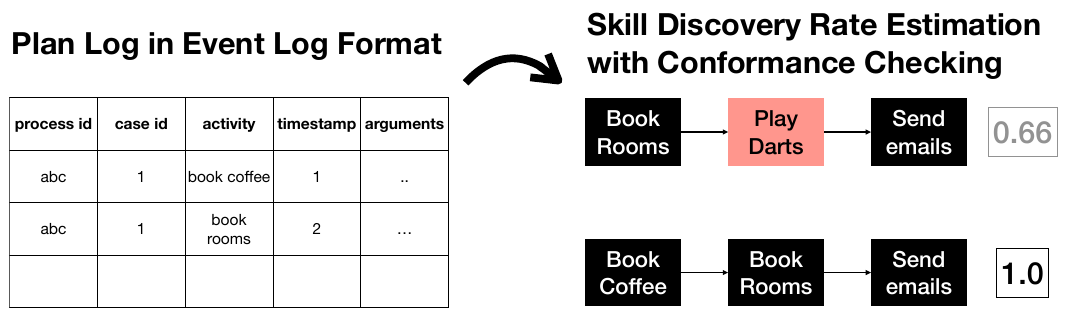}
        \caption{Fig.~\thesubfigure\ Schematic view of the experimental pipeline for skill learning with process discovery.}
        \vspace*{0.4cm}
        \label{fig-3a}
    \end{subfigure}
    \par
    \begin{subfigure}[t]{\linewidth}
        \centering
        \includegraphics[width=\linewidth]{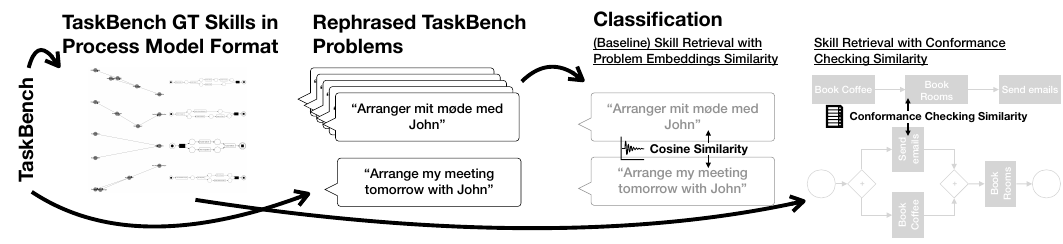}
        \caption{Fig.~\thesubfigure\ Schematic view of the experimental skill retrieval pipeline with conformance checking.}
        %\vspace*{-1cm}
        \label{fig-3b}
    \end{subfigure}
\end{figure}

\vspace{-0.6cm}
\subsubsection{Skill Learning With Process Discovery} 
To evaluate the feasibility and effectiveness of our proposed skill learning method, we conducted experiments using the ProcessTBench synthetic dataset \cite{redis_processtbench_2024}. This experiment aimed to determine whether our method could reliably generate accurate process models (skills). We used conformance checking of the generated action sequences (traces) compared to ground-truth process models provided in the dataset.

The ProcessTBench synthetic dataset includes queries requiring LLM-generated plans, with corresponding solutions provided as action sequences using predefined tools. Each of the 532 problems represents a distinct process instance, with 5-6 LLM-generated action sequences serving as individual cases within these instances. The dataset simulates multiple process executions, where different paraphrasings of a query and various plans are treated as execution instances of the same process.

To evaluate the feasibility of our skill learning method, we used conformance checking to compare the LLM-generated plans (traces) from ProcessTBench with the ground truth process models provided by TaskBench. This part is illustrated in Fig.~\ref{fig-3a}. We employed two widely-used conformance checking metrics, \textit{Replay Fitness} and \textit{Alignment Fitness}~\cite{jagadeesh2010trace}, to assess how well the generated plans align with the known process models. High conformance values indicate that accurate process models could be derived from these traces using process discovery, supporting our method's viability. For this task, ProcessTBench uses the \texttt{GPT-4-0613} model.

\vspace{-0.2cm}
\subsubsection{Skill Retrieval With Conformance Checking} We conducted experiments using the TaskBench dataset, rephrasing 533 problems (queries) 5-6 times each to simulate scenarios where a similar problem requires a relevant solution from the skill library. We then tested different skill retrieval methods. We compared the baseline method that uses problem embeddings, our proposed model using conformance checking, and a hybrid model combining both approaches (i.e., two-stage retrieval). We aimed to determine if conformance checking could enhance retrieval accuracy compared to established similarity methods.

The architecture of the Skill Retrieval With Conformance Checking experiment is depicted in Fig.~\ref{fig-3b}. This architecture has the following components.
\begin{enumerate}
    \item \textbf{LLM Rephraser}: Generates paraphrased descriptions of problems in English, Danish, and French. To improve its accuracy, the model was asked first to reason about the problem, as shown in~\cite{mizrahi_state_2023}.
        \begin{description}
            \item Input: Problem description from the TaskBench dataset~\cite{shen_taskbench_2023}
            \item Output: Rephrased problem descriptions
        \end{description}
    \item \textbf{DAG to Petri net Converter}: Converts reference DAG process models to Petri nets.
        \begin{description}
            \item Input: A reference process model presented in the DAG format
            \item Output: A process model presented in the Petri net format
        \end{description}
    \item \textbf{LLM plan (\textit{thought}) generator}: Generates partial plans (\textit{thoughts}) using the given problem description and tools available in the TaskBench domain. We call these plans \textit{thoughts} and partial to distinguish them from the output of more thorough and widely used planners such as ReAct \cite{yao_react_2023}. I.e., due to experimental constraints, the thought generator is asked to return a full trace solution in one inference session, while ReAct builds the plan more thoroughly in inference iterations.
        \begin{description}
            \item Input: Problem description and all available tools in the TaskBench domain
            \item Output: Plan that solves the problem as a sequence of actions.
        \end{description}
    \item \textbf{Classifier}: Calculates a distance distribution between rephrased and original TaskBench problems. For conformance checking retrieval, we first generate a thought—a partial plan representing the solution before calculating the distance. This thought is then used to compare the rephrased problem to other problems.
        \begin{description}
            \item Input: Rephrased problem
            \item Output: Distance distribution over original problems.
        \end{description}
\end{enumerate}

For the classifier, the following two baseline and proposed models were used:
\begin{enumerate}
    \item \textbf{Universal Sentence Encoder (USE):} Measures semantic similarity using cosine distance between problem and skill embeddings~\cite{cer_universal_2018}.
    \item \textbf{ada-002:} A more recent embedding model by OpenAI~\footnote{https://openai.com/blog/new-and-improved-embedding-model}.
    \item \textbf{Conformance checking:} Measures the alignment between the LLM-generated plan (\textit{thought}) and the stored process models, using alignment fitness as a similarity measure~\cite{jagadeesh2010trace}.
    \item \textbf{Hybrid (ada-002 + Conformance checking):} Candidates are pre-filtered using ada-002 and re-ranked using conformance checking.
\end{enumerate}

The retrieval models' performance was evaluated using two state-of-the-art metrics for multi-class classification and next-item recommender systems: F1-score and Mean Reciprocal Rank (MRR)~\cite{taha_metrics_2015}.

Each of these evaluation metrics serves different purposes. For a given rephrased query, the models output a ranked list of process models based on their similarity scores. The F1-score measures the accuracy when considering only the top-ranked prediction, counting it as correct if it matches the actual class. The Mean Reciprocal Rank (MRR) provides a more nuanced assessment by evaluating the average inverse rank position of the correct class in the ranked list, thus accounting for how high the correct class appears in the recommendations.

\subsection{Skill Learning with Process Discovery}

In this study, we evaluated the feasibility of our proposed skill learning method using the TaskBench~\cite{shen_taskbench_2023} and the ProcessTBench \cite{redis_processtbench_2024} synthetic datasets. TaskBench contains various queries and their corresponding process models. Each query represents a separate process, and each specific instance of solving that query is considered a case. ProcessTBench extends TaskBench by providing repeated sequential planning traces for the problems.

Table~\ref{tab-learning} summarizes the dataset characteristics, including the average number of cases per process, the number of activities (actions) per process, and the fitness metrics for conformance checking. The mean number of cases per process was 4.08, indicating a relatively small dataset, especially compared to popular real-world process mining datasets like the BPI Challenges, where the number of cases is significantly larger~\cite{sani_prototype_2019}. Despite the dataset's small size, the high average replay fitness (0.96) and alignment fitness (0.94) suggest that our skill learning method is feasible. These metrics indicate that the discovered skills closely matched the ground truth in many cases, with 75\% processes achieving a perfect fitness score for all associated cases. 

\begin{table}[tb]
    \centering
    \caption{Descriptive statistics of the dataset used to evaluate the feasibility of skill acquisition through process discovery.} 
    \label{tab-learning}
    \begin{tabular}{lrrrrr}
    \toprule
     & Mean& Std.& Min.& Median& Max.\\
    \midrule
    Cases (Plans) / Process& 4.08 & 1.27 & 2 & 4 & 11 \\
 Activities (Actions) / Process& 3.79 & 0.89 & 2 & 4 &8 \\
    Case Variants / Process& 2.68 & 0.97 & 1 & 3 & 5 \\
    Replay Fitness / Case& 0.96 & 0.09 & 0.25 & 1.00 & 1.00\\
    Alignments Fitness / Case& 0.94 & 0.12 & 0.40 & 1.00 & 1.00 \\
    \bottomrule
    \end{tabular}

\end{table}

\begin{figure}[h]
    \centering
    \includegraphics[width=0.4\linewidth]{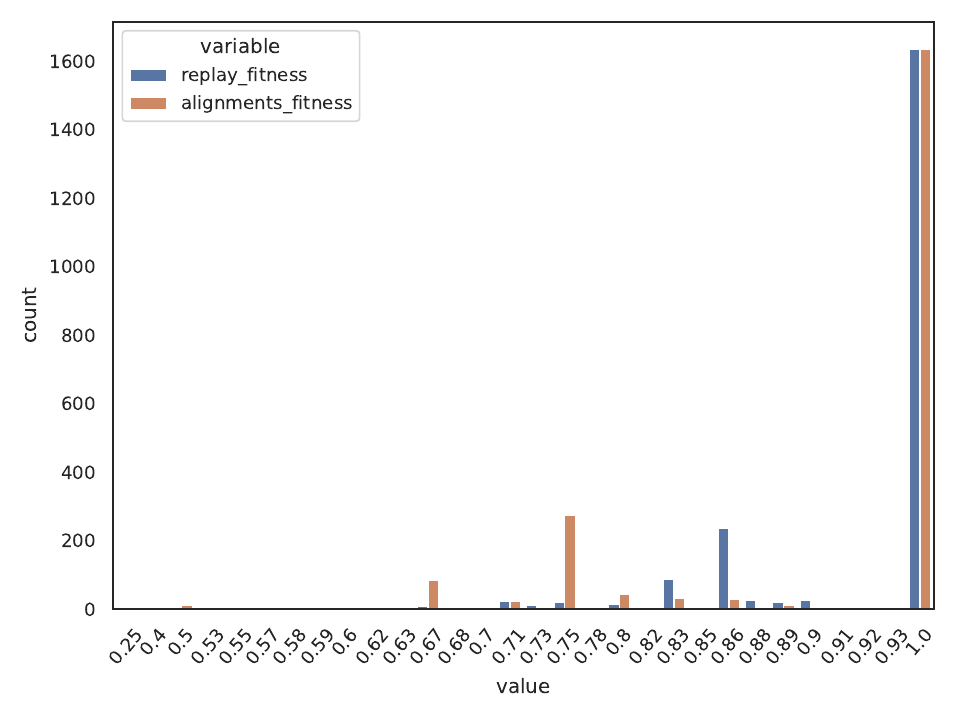}
    \caption{Summary of the overall accuracy of the traces generated by the planner for n=2173 cases.}
    \label{fig-planstoredistr}
\end{figure}

Figure~\ref{fig-planstoredistr} illustrates the accuracy distribution of the generated traces. Replay fitness tends to be higher than alignments, suggesting the planner's difficulty assembling individual components into a coherent sequence. E.g., for a specific query in ProcessTBench, the ground truth process model was  \(\wedge(\rightarrow(A,B),\rightarrow(C,D,E,F)) \), but the planner produced the sequence  E, F, A, B, C, D, resulting in a replay fitness of 0.9 and an alignment fitness of 0.66. Although the planner accurately captured most relationships between actions, it struggled with correctly ordering them. This suggests that sequence alignment, and consequently the creation of complete control flow models, could be particularly challenging for LLMs — a challenge that is somewhat expected given that reasoning is a known weakness. 

%Furthermore, this could suggest that LLMs alone may struggle to reliably create complete control flow models for processes and their associated actions. Such models are crucial for leveraging parallelism, as they allow for identifying independent actions that can be executed concurrently, thereby enhancing efficiency and performance in complex tasks. 

The experiment aimed to demonstrate that a real-life dataset of 'similar problems' would exhibit sufficient variance to allow the capture of accurate process models. However, in this study, the variance was artificially generated using the synthetic ProcessTBench dataset, where an LLM was prompted to 'generate more plan variants,' which may not fully replicate real-world conditions. Finally, although ProcessTBench contains a relatively low number of traces, process discovery algorithms are not the sole real-life contributors to skill learning. Human experts can also play a crucial role in defining the correct skills. Overall, the evidence suggests that when automated process discovery tools are combined with additional data and expert input, it is likely feasible to construct a robust skill library using our proposed skill learning method.

\subsection{Skill Retrieval with Conformance Checking}
The results in Table~\ref{tab-search-overall} show that the ada-002 model outperforms the other models in terms of F1 score and MRR. The combination of ada-002 and Conformance Checking also delivers strong performance, nearly matching the accuracy of the ada-002 model. This suggests at first glance that the baseline method, using ada-002, is better for Skill Retrieval than our proposed methods using conformance checking.

\begin{table}[tb]
    \caption{F1 score and MRR for the four retrieval methods. The best result across each metric is highlighted in bold. The ada-002 X Conformance Checking @ 3 is the hybrid model described in Section~\ref{sec-experiments}, with 3 nearest neighbors retrieved during the first stage.}
    \label{tab-search-overall}
    \centering
    \begin{tabular}{lrr}
    \toprule
     & F1 & MRR \\
    \midrule
    USE & 0.66 & 0.71 \\
    ada-002 & \textbf{0.91} & \textbf{0.94} \\
    Conformance Checking & 0.75 & 0.84 \\
    ada-002 X Conformance Checking @ 3 & 0.90 & 0.93 \\
    \bottomrule
    \end{tabular}
    \vspace{-1em}
\end{table}

It is worth observing that ada-002 performs 11\% better than in the documentation provided by OpenaAI, the model's creators. This suggests that our rephrased queries were less complex than what was used in their benchmark (sentence similarity); see Footnote 2. We couldn't find previous work on a comparable experiment for conformance checking.

\begin{figure}[tb]
    \centering
    \includegraphics[width=\linewidth]{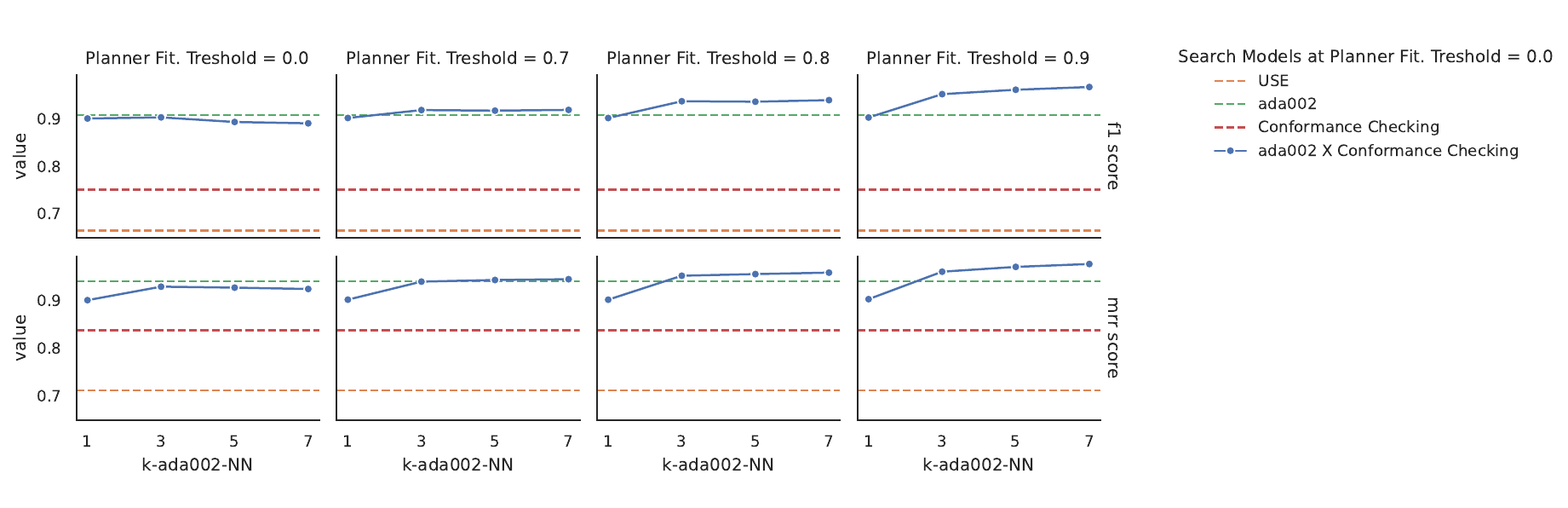}
    \vspace{-0.6cm}
    \caption{performance of the proposed ada-002 and conformance hybrid model, given different planner accuracies. The columns represent different planner accuracies, while the rows represent the metric, f1, or MRR. The x-axis represents the nearest neighbors retrieved by the first stage retrieval model, ada-002. The y-axis represents the final result of the respective metric after reranking with Conformance Checking. The reference lines show the results at the baseline planner accuracy (fitness) threshold = 0.0. The figure reveals that with a counterfactual conformance checking threshold of 0.7 alignment fitness, the combined ada-002 and Conformance Checking would outperform all other methods.}
    \label{fig-search-sensitivity}
    \vspace{-0.2cm}
\end{figure}

Finally, we show that the results above might change when relaxing one experimental constraint: the accuracy of the planner. To understand the impact of the plan (\textit{thought}) generator accuracy on the relative performance of our models, we did a sensitivity analysis, shown in Fig.~\ref{fig-search-sensitivity}. It turns out that the combined ada-002 and conformance-checking model would outperform ada-002, given a generator accuracy of 0.7 or more. In our opinion, a threshold of 0.7 is not too low. It could be realistically achieved with better planner design (cp. with the planner accuracy~\ref{tab-learning} with an average of 0.94 and std of 0.12, and with TaskBench achieving 0.9 node and 0.71 edge prediction~\cite{shen_taskbench_2023}). 

These findings lead us to believe that, with better "thought" generator design, integrating conformance checking with other retrieval models could be a promising direction for future research. As discussed in Section~\ref{sec-methods-retrieval}, the proposed Skill Retrieval approach is also more interpretable and accurate.

While generating 'thoughts' initially introduces additional LLM inference overhead, this investment can lead to overall inference savings compared to methods like ReAct. Since the generated plans can be stored and reused for similar problems in the future, the need for repeated inference is reduced. Additionally, the ability to identify and execute parallelizable actions within these plans allows for concurrency savings, further enhancing efficiency.

Finally, it's important to note that ada-002 has been outperformed by newer embedding models since conducting these experiments, with improvements of up to 25\%~\cite{muennighoff_mteb_2022}. Similarly, more recent LLMs have surpassed the one used for skill retrieval in this study. These advancements could significantly impact the results, suggesting that future research should evaluate the effects of these newer models on skill learning and retrieval.

%%%%%%%%%%%%%%%%%%%%%%%%%%%%%%%%%%%%%%%%%%%%%%%%%%%%%%%
\section{Conclusion}

Our experiments suggest the feasibility of using process mining techniques for skill learning in text-based LLM planners. By integrating conformance checking into skill retrieval, we may improve the accuracy and interpretability of plan generation. These advancements could pave the way for more efficient and reliable LLM-driven automation solutions. More interpretable LLM plan generation can give businesses greater control and transparency over automated processes. As LLMs' role in decision-making expands, effectively managing and understanding these systems will be increasingly important. 

In the future, we aim to develop our experimental framework further. Generating more data and creating more sophisticated LLM plan generators will enable more general conclusions. Furthermore, controlling for more confounding variables, such as LLM and embedding models, would increase the robustness of the observed effects. We have developed a skill learning framework and presented its theoretical properties. Still, we have not created a skill "usage" framework to gather evidence for our approach's end-to-end effectiveness. Lastly, we have tested skill retrieval in a closed-set classification setting, but the open-set classification is also a realistic use case worth exploring. 
%%%%%%%%%%%%%%%%%%%%%%%%%%%%%%%%%%%%%%%%%%%%%%%%%%%%%%%

\bibliographystyle{splncs04}
\bibliography{references}   
\end{document}